\documentclass[runningheads]{llncs}
\usepackage{graphicx}
\usepackage{amsfonts}       
\usepackage{mathrsfs}
\usepackage{amsmath,amssymb}
\usepackage{tikz}
\usepackage{soul}
\usepackage{breakcites}
\usepackage[final]{pdfpages}
\usepackage{hyperref}       

\newcommand{\beginsupplement}{%
    \setcounter{table}{0}
    \renewcommand{\thetable}{S\arabic{table}}%
    \setcounter{figure}{0}
    \renewcommand{\thefigure}{S\arabic{figure}}%
    \setcounter{section}{0}
    \renewcommand\thesection{\Alph{section}}%
}

\begin{document}
\title{Region Proposals for Saliency Map Refinement for Weakly-supervised Disease Localisation and Classification\thanks{Supported by Australian Research Council through grants DP180103232}}
%
%
\author{Renato Hermoza$^{\dagger}$ \qquad Gabriel Maicas$^{\dagger}$ \qquad  Jacinto C. Nascimento$^{\ddagger}$ \qquad Gustavo Carneiro$^{\dagger}$}
%
%
\institute {$^{\dagger}$Australian Institute for Machine Learning, The University of  Adelaide \\ $^{\ddagger}$Institute for Systems and Robotics, Instituto Superior Tecnico, Portugal}
%
\maketitle              
\begin{abstract}

The deployment of automated systems to diagnose diseases from medical images is challenged by the requirement to localise the diagnosed diseases to justify or explain the classification decision. 
This requirement is hard to fulfil because most of the training sets available to develop these systems only contain global annotations, making the localisation of diseases a weakly supervised approach. 
The main methods designed for weakly supervised disease classification and localisation rely on saliency or attention maps that are not specifically trained for localisation, or on region proposals that can not be refined to produce accurate detections. 
In this paper, we introduce a new model that combines region proposal and saliency detection to overcome both limitations for weakly supervised disease classification and localisation. 
Using the ChestX-ray14 data set, we show that our proposed model establishes the new state-of-the-art for weakly-supervised disease diagnosis and localisation.
\end{abstract}
\section{Introduction}
\label{sec:intro}

An important way to explain a disease classification made by a medical image computing (MIC) system relies on showing the image region(s) associated with the classification.
Even though object detection is a classic MIC problem, it usually relies on the availability of fully annotated data sets that contain not only the disease classification, but also the localisation of image regions associated with the classification~\cite{litjens2017survey}.
Unfortunately, such data sets tend to be expensive to acquire and small, which is challenging for the training of classification and detection models, particularly the ones based on deep learning.
Such issues motivated the community to consider data sets that are larger but weakly supervised~\cite{wang_chestx-ray8_2017}.
Since the deployability of disease diagnosing systems partly depends on the localisation of image regions associated with the image classification, the medical image analysis community is increasingly developing systems that can classify and localise diseases from  weakly annotated training sets~\cite{guan_thorax_2020,tang_weakly_2018,wang_chestx-ray8_2017,yao_weakly_2018}.

Currently, weakly-supervised disease detection and classification methods merge the classification model with saliency maps~\cite{wang_chestx-ray8_2017,yao_weakly_2018}, region proposal~\cite{li_thoracic_2018,wang2019weakly}, or attention maps~\cite{guan_thorax_2020,liu_align_2019,tang_attention-guided_2018}.
Methods based on saliency maps~\cite{wang_chestx-ray8_2017,yao_weakly_2018} represent the most common approach in the field. 
Saliency map methods produce classification results based on a pooling operation from the model's last layer that contains an activation pattern that is likely to highlight image regions that are active for the classification result. However, there is no penalisation when the saliency map highlights image regions that are not associated with the classification of a particular disease. 
Region proposal methods explicitly encourage regions to be associated with correct classification labels and penalise regions associated with incorrect classification labels~\cite{li_thoracic_2018}. 
However, they do not have a way to differentiabily extract particular crops (or regions), so they aggregate the information from all extracted crops~\cite{bilen_weakly_2016,li_thoracic_2018}. 
Consequently, they are unable to refine the initially proposed regions as done by supervised methods~\cite{he_mask_2017,ren_faster_2015}.
Attention maps extend saliency maps by enforcing the classification of a highlighted image region~\cite{guan_thorax_2020,liu_align_2019,tang_attention-guided_2018} and penalising regions associated with incorrect classification~\cite{maicas_model_2019}. 
However, comparison with these methods is difficult given that they use quantitative evaluation measures that cannot be used for a fair comparison with other approaches~\cite{tang_attention-guided_2018} and they also use unpublished data set splits~\cite{guan_thorax_2020,liu_align_2019} for ChestX-ray 14 data set~\cite{wang_chestx-ray8_2017}.

In this paper, we introduce a novel model that jointly produces disease diagnosis and localisation of relevant image regions associated with the diagnosis, where the localisation process relies on combining the results from the region proposal and saliency map detectors. 
Such combination is enabled by the use of the Gumbel softmax function~\cite{jang_categorical_2017} to differentiablily sample discrete regions from a set of region proposals, which allows us to refine the proposed regions and potentially increase the weakly-supervised detection precision. 
We test our new approach on the ChestX-ray14 data set~\cite{wang_chestx-ray8_2017} using the published train-test split and widely used quantitative evaluation measures. 
Results show that we establish a new state-of-the-art for both classification, with 0.82 average area under the receiver operating characteristic curve (AUC), weakly supervised localisation results with 0.29 average intersection over union (IoU) and 0.37 average continuous Dice (cDice).
We will make our code publicly available (upon acceptance of our paper) to foster reproducibility and trustworthy research on the field.

\section{Related Works}
\label{sec:rel}

Weakly supervised disease classification and localisation is gaining increasing attention by the MIC community. This is partly due to the availability of relatively large chest X-ray data sets designed to be used in the development of models that can address weakly supervised disease classification and localisation~\cite{irvin_chexpert_2019,wang_chestx-ray8_2017}.  
Chest X-ray imaging is one of the most widely available modalities for screening and diagnosis. However, automatic disease classification and localisation from chest X-ray images is recognised as being technically challenging~\cite{folio2012chest}. 
This is primarily due to (i) the large diversity in the appearance, size and location of the visual patterns of different types of thoracic diseases, and (ii) the relatively scarcity of high quality disease annotations. 
The initial models proposed for this problem were based on standard deep learning models~\cite{rajpurkar_chexnet_2017,wang_chestx-ray8_2017} that produced relatively accurate classification, but poor detection results.
More recent approaches improve classification accuracy by handling label noise~\cite{irvin_chexpert_2019}, incorporating information from the associated radiology reports into the training process~\cite{chen2019deep,tang_attention-guided_2018}, or including strong annotations to add extra supervision~\cite{guendel_learning_2018,li_thoracic_2018,liu_align_2019}. 

Research on diagnosing and localising diseases from chest x-rays (especially based on ChestX-ray14)~\cite{wang_chestx-ray8_2017} has been hindered by the following two factors: 1) the data set split proposed by~\cite{wang_chestx-ray8_2017} for the evaluation is not often used, and 2) the localisation results are reported using different evaluation measures.
These issues make a fair comparison between different approaches challenging.
Regarding data set splits, some  approaches~\cite{guan_multi-label_2018,guendel_learning_2018,ma_multi-label_2019,tang_attention-guided_2018,wang_thorax-net_2019,wang_chestx-ray8_2017,yao_weakly_2018} use the  published split~\cite{wang_chestx-ray8_2017}, while others~\cite{guan_thorax_2020,li_thoracic_2018,liu_align_2019,rajpurkar_chexnet_2017,yao_learning_2017} use a random split, which is not appropriate because it leads to unfair comparisons, as mentioned above -- for example, it is possible that images from the same patient can be present in both training and testing sets in these random splits. 
This is discussed by Wang~\textit{et al.}~\cite{wang_thorax-net_2019} by showing that results on the suggested split~\cite{wang_chestx-ray8_2017} can be worse than in random splits by more than 10\% average AUC. 
Regarding localisation measures, most methods~\cite{li_thoracic_2018,liu_align_2019,wang_chestx-ray8_2017} use IoU (but
Li \emph{et al.}~\cite{li_thoracic_2018} uses a few bounding box annotations for training and Liu \emph{et al.}~\cite{liu_align_2019} relies on an unpublished train/test split and also uses a few bounding box annotations for training), while~\cite{yao_weakly_2018} uses cDice.
While IoU is a standard measure for object detection, it is sensitive to the threshold applied to binarise the saliency map~\cite{yao_weakly_2018}.
This issue is alleviated by the cDice measure that does not binarise the detection map -- instead it is based on the continuous values of the saliency map.
In this paper, we use the experimental setup proposed by Wang~\textit{et al.}~\cite{wang_chestx-ray8_2017} to ensure that our results are fairly compared with previous methods and can be used as baseline for future approaches.

\section{Method}
\label{sec:method}

\begin{figure}[t]
\centering
\includegraphics[width=1.00\linewidth]{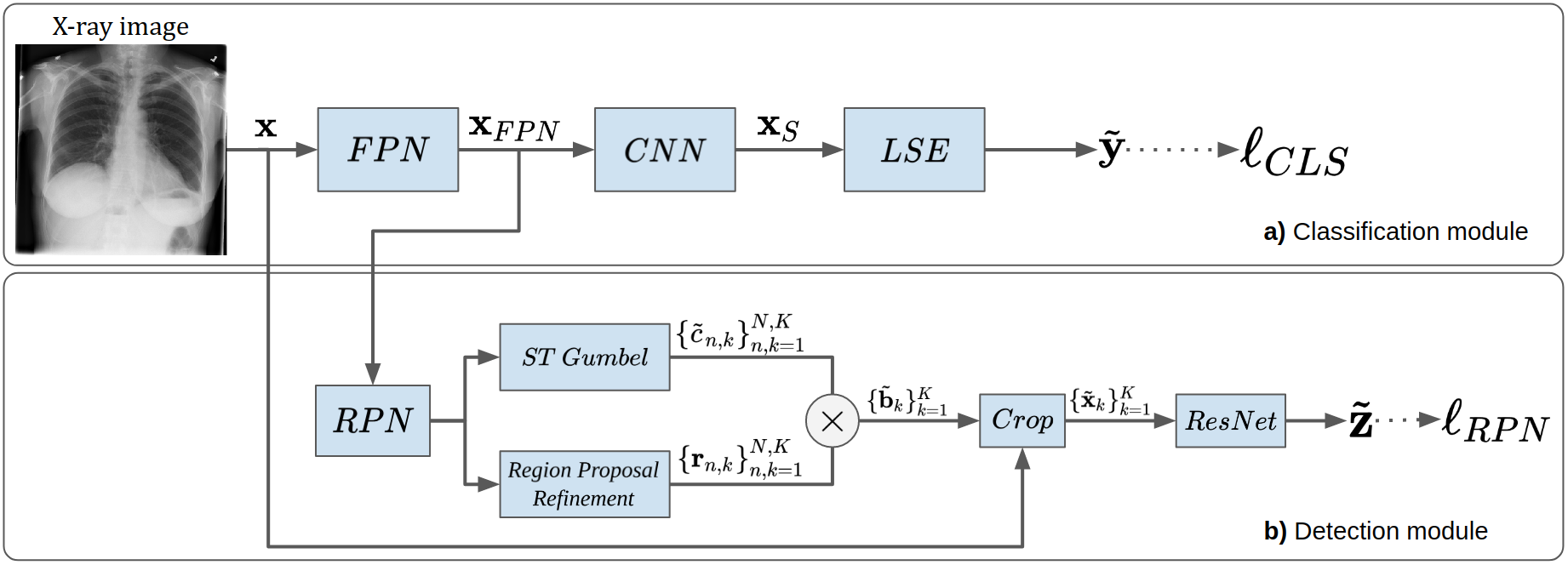}
\caption{
The architecture of the proposed model consists of two modules:
a) a classification module that produces saliency map $\mathbf{x}_S$ (via class activation maps) and classification $\tilde{\mathbf{y}}$ (trained with loss $\ell_{CLS}$ in Eq.~\ref{eq:l_cls});
and b) a detection module that produces a set of regions $\{ \mathbf{r}_{n,k} \}_{n,k=1}^{N,K}$ with class confidences $ \{ c_{n,k} \}_{n,k=1}^{N,K}$ and region proposal classification $\tilde{\mathbf{z}}$ (trained with loss $\ell_{RPN}$ in Eq.~\ref{eq:l_rpn}).
}
\label{fig:model}
\end{figure}

The proposed weakly supervised disease classification and detection consists of a joint classification and detection approaches, where the detection combines the results from saliency map and region proposal, as shown in Fig.~\ref{fig:model}. We first explain the training and testing sets used, followed by an explanation of the model inference and training approaches.

\begin{figure}[t]
\centering
\includegraphics[width=0.85\linewidth]{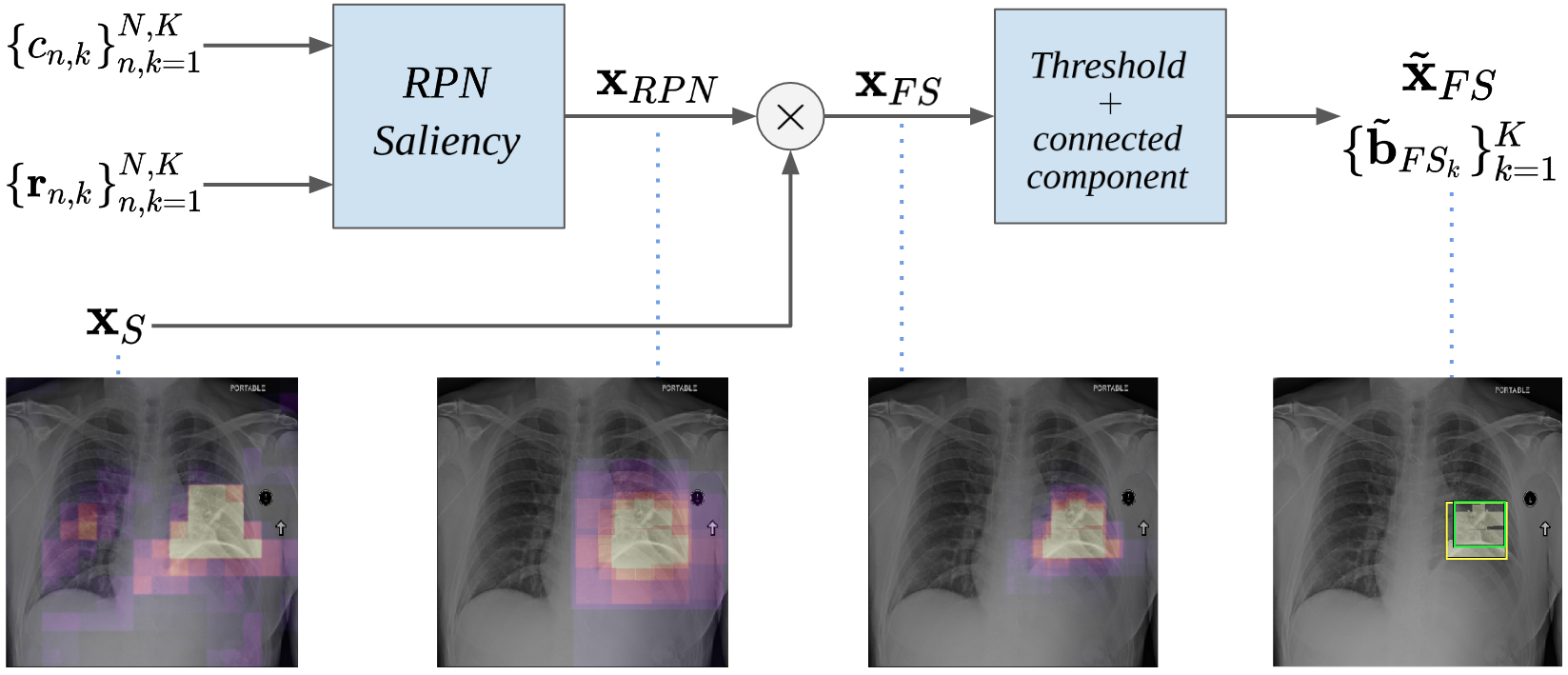}
\caption{
Inference procedure to obtain the final bounding boxes from saliency maps.
Sources of saliency maps from left to right:
1) classification module $\mathbf{x}_S$;
2) ROI detection scores $\mathbf{x}_{RPN}$;
3) saliency map combination $\mathbf{x}_{FS} = \mathbf{x}_S \odot \mathbf{x}_{RPN}$
4) binarised saliency $\tilde{\mathbf{x}}_{FS}$ with  bounding box $\tilde{\mathbf{b}}_{FS_k}$ (predicted bounding box is shown in yellow and ground truth in green).
}
\label{fig:inference}
\end{figure}

\subsection{Data Set}

The training set is defined by $\mathcal{D} = \{(\mathbf{x},\mathbf{y})_i\}_{i=1}^{|\mathcal{D}|}$ and the testing set is formed by $\mathcal{T} = \{(\mathbf{x},\mathbf{y},\{ \mathbf{b}_k \}_{k=1}^K)_i\}_{i=1}^{|\mathcal{T}|}$, where $\mathbf{x}: \Omega \rightarrow \mathbb R$ denotes an X-ray image, with $\Omega$ being the image lattice, $\mathbf{y} \in \{0,1\}^{K}$ indicates the presence or absence of $K$ pathologies, and $\mathbf{b}_k \in \mathbb R^{4}$ indicates the bounding box (center coordinates, width and height) localising each of the $K$ pathologies (note that if the $k^{th}$ pathology is not present in a test image or if it was not annotated, the $k^{th}$ element of the set $\{ \mathbf{b}_k \}_{k=1}^K$ contains a token indicating that the annotation is not available). 

\subsection{Weakly Supervised Disease Classification and Detection}
\label{sec:model}

The system integrates a classification and a detection modules -- see Fig.~\ref{fig:model}.
The classification module follows a fully convolutional model~\cite{oquab_is_2015}, consisting of a feature pyramid network (FPN)~\cite{lin_feature_2017} that extracts $\mathbf{x}_{FPN} \in \mathbb R^{Q \times (H_{x}/4) \times (W_{x}/4)}$, which contains $Q$ feature maps of size $(H_{x}/4) \times (W_{x}/4)$ (with $H_x \times W_x$ representing the height and width of image $\mathbf{x}$), followed by a convolutional neural network (CNN) that produces $K$ saliency maps of size $(H_{x}/4) \times (W_{x}/4)$ (one map for each of the $K$ classes), denoted by $\mathbf{x}_{S} \in \mathbb R^{K \times (H_{x}/4) \times (W_{x}/4)}$, and the final classification logit $\tilde{\mathbf{y}} \in \mathbb R^K$ is produced by pooling the results from each of the saliency maps using the log-sum-exp (LSE) function. 

The detection module takes $\mathbf{x}_{FPN}$ and uses $RoiAlign$~\cite{he_mask_2017} to extract the features of $N$ pre-defined region proposals for each of the $K$ classes, where each region proposal is defined by a 4-dimensional bounding box vector.  The features from these $N \times K$ region proposals are used by two regressors: one to predict the class confidence for each region proposal $ \{ c_{n,k} \}_{n,k=1}^{N,K}$, and another to predict the refined region proposal bounding box vector $\{ \mathbf{r}_{n,k} \}_{n,k=1}^{N,K}$, with $\mathbf{r}_{n,k} \in \mathbb R^4$. Instead of aggregating all region proposals that can produce inaccurate detection results because of the large value of $N$, we use the differentiable operator Straight-Through (ST) Gumbel-Softmax estimator~\cite{jang_categorical_2017} to sample a single region proposal per class based on the confidence value -- effectively, the result of this operator forms a binarised $c_{n,k}$, denoted by 
$\tilde{c}_{n,k} \in \{0,1\}$, where $\sum_{n=1}^N \tilde{c}_{n,k}=1$ for each class $k$. 
By selecting the region proposals $\mathbf{r}_{n,k}$ for which $\tilde{c}_{n,k}=1$ we build the bounding box set $\{ \tilde{\mathbf{b}}_k \}_{k=1}^K$ (with $\tilde{\mathbf{b}}_k \in \mathbb R^4$), which is used to crop the input image $\mathbf{x}$ to produce $K$ feature maps $\{ \tilde{\mathbf{x}}_k \}_{k=1}^K$ (where $\tilde{\mathbf{x}}_k \in \mathbb R^{3 \times H_f \times W_f}$, with $H_f \times W_f$ denoting the height and width of the crop operation) that are used by a ResNet~\cite{he_deep_2016} to produce a region proposal classification denoted by the logit $\tilde{\mathbf{z}} \in \mathbb R^K$.

The training procedure minimises the binary cross entropy loss for each class $k$ for the classification and the detection modules.  Due to the high number of cases with no pathologies, we adopt a balancing strategy for the labels using positive and negatives weight factors $\beta_P,\beta_N \in \mathbb R^K$.
The loss for the classification module for each sample $i$ is:
\begin{equation}
    \ell_{CLS}(\tilde{\mathbf{y}}_i,\mathbf{y}_i) = -\beta_P\mathbf{y}_i\log{(\tilde{\mathbf{y}}_i)}
    -\beta_N(1-\mathbf{y}_i)\log{(1-\tilde{\mathbf{y}}_i)}
    \label{eq:l_cls}
\end{equation}
where $\beta_P(k) = 1 - \frac{P_k}{|\mathcal{D}|}$, $\beta_N(k) = \frac{P_k}{|\mathcal{D}|}$, where $P_k$ is the total number of positive cases for class $k$ and $|\mathcal{D}|$ is the training set size.
In a similar way, we define the following loss for the detection module:
\begin{equation}
    \ell_{RPN}(\tilde{\mathbf{z}}_i,\mathbf{y}_i) = -\beta_P\mathbf{y}_i\log{(\tilde{\mathbf{z}}_i)}
    -\beta_N(1-\mathbf{y}_i)\log{(1-\tilde{\mathbf{z}}_i)}.
    \label{eq:l_rpn}
\end{equation}
The training of the detection module is sensitive because region proposals are unstable at the beginning of the optimisation.
To address this issue, we divide the training process into 3 stages:
1) training of the classification module with $\ell_{CLS}$ in~\eqref{eq:l_cls},
2) training of the detection module with $\ell_{RPN}$ in~\eqref{eq:l_rpn}, and
3) joint training of the classification and detection modules with loss $\ell_{CLS} + \ell_{RPN}$.

The inference procedure combines the saliency map and region proposal detection results, by:
1) computing the average of all refined region proposals weighted by their confidence, denoted by $\mathbf{x}_{RPN} \in \mathbb R^{K \times (H_x/4) \times (W_x/4) }$;
2) followed by an element-wise multiplication with the saliency map from the classification module, defined by $\mathbf{x}_{FS} = \mathbf{x}_S \odot \mathbf{x}_{RPN}$;
3) thresholding followed by the selection of the largest connected component which produces the final binary segmentation map $\tilde{\mathbf{x}}_{FS} \in \{0,1\}^{K \times (H_x/4) \times (W_x/4)}$; and
4) obtaining the parameters of a bounding box $\{ \tilde{\mathbf{b}}_{FS_k} \}_{k=1}^K$ (with $\tilde{\mathbf{b}}_{FS_k} \in \mathbb R^4$) as the smallest rectangle able to cover the pixels $\omega \in \Omega$ where $\tilde{\mathbf{x}}_{FS}(\omega) = 1$.
The inference procedure is shown in Fig.~\ref{fig:inference}.

\section{Experiments}
\label{sec:exp}

\begin{figure}[t]
\centering
\includegraphics[width=1.00\linewidth]{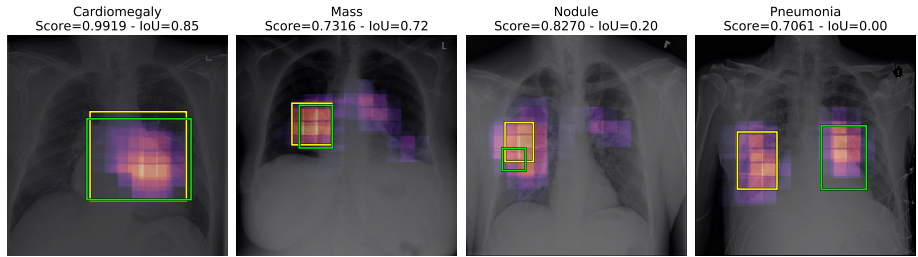}
\caption{Examples of classification and detection results. Predicted bounding boxes are shown in yellow and ground truth in green.
}
\label{fig:samples}
\end{figure}

\subsection{Data Set}

We conduct the experiments in this paper on the data set ChestX-ray14~\cite{wang_chestx-ray8_2017}, which contains 112,120 frontal-view chest X-ray images that are weakly labelled for 14 pathologies (this is a multi-label problem, where each image can have between 0 and 14 annotated pathologies) and bounding box annotations for 880 images relative to 8 different diseases.
We use the  published training and test split provided with the data set~\cite{wang_chestx-ray8_2017}, and we use part of the training set as validation for model selection (i.e., hyper-parameter estimation).
Note that the bounding box annotations are only present in some of the testing set images, i.e. they are not used for training.

\subsection{Experimental Set Up}

We train the model in three stages (as described in Sec.~\ref{sec:model}), where in stages 1 and 2 we use a learning rate of 0.001 and for stage 3 a learning rate of 0.0003. We use Adam~\cite{kingma_adam:_2015} with a momentum of 0.9, weight decay of 0.001 and a mini-batch size of 8. 
Images are down-sampled to 512 x 512 and normalised using ImageNet~\cite{russakovsky_imagenet_2015} mean and standard deviation. 
While training, we apply random data augmentation operations such as: zoom between 0 and 0.1, translation in the four directions between -50 and 50 pixels, rotation between -10 and 10 degrees and random horizontal flipping. 
We use ImageNet~\cite{russakovsky_imagenet_2015} pre-trained Densenet-121~\cite{huang_densely_2017} and ResNet-34~\cite{he_deep_2016} models to initialise $CNN$ and $ResNet$ respectively (see Fig.~\ref{fig:model}),
where we replaced all ReLU activations with Leaky ReLU~\cite{maas_rectifier_2013}, using a 0.1 negative slope.
The initial region proposals in RPN uses regions of size 64, 128 and 256, all with ratio 1 and strides of 8, 16 and 32, to produce a total of $N=395$ region proposals.
The ST Gumbel-Softmax estimator temperature $\tau$ is exponentially annealed from 1 to 0.001 on stages 2 of 3 of the training procedure.
The thresholds used to obtain the predicted bounding boxes are empirically selected for each class using the ground truth annotations\footnote{This practice follows the protocol of other methods~\cite{guan_thorax_2020,wang_chestx-ray8_2017} in the field.}.

To evaluate our model, we use the AUC for each pathology and the average over pathologies.
For detection accuracy, we evaluate in terms of average IoU and cDice.
We compared our method against baselines methods using saliency maps~\cite{wang_chestx-ray8_2017,yao_weakly_2018}, region proposals~\cite{li_thoracic_2018} and attention maps~\cite{guan_multi-label_2018,ma_multi-label_2019}.
Note that to allow a fair comparison, we only include methods that used the  published train-test split~\cite{wang_chestx-ray8_2017} and the widely used detection measures IoU and cDice.

\subsection{Results}

Table~\ref{tab:res1} compares the AUC classification results between our approach (labelled as 'Ours') and several baselines~\cite{guan_multi-label_2018,li_thoracic_2018,ma_multi-label_2019,wang_chestx-ray8_2017}.
We show an ablation study for the detection results of our method in Table~\ref{tab:comparison} using different saliency maps to select the final bounding box:
the classification module saliency map $\mathbf{x}_S$ (denoted by Sal), the region proposal map $\mathbf{x}_{RPN}$ (denoted by Det), and the combined saliency and region proposal maps $\tilde{\mathbf{x}}_{FS}$ (denoted by Mix).
As activation maps tends to highlight more regions, we can observe that saliency map Sal performs well on Cardiomegaly (the largest pathology on the dataset) while saliency map Det performs better on the rest of labels.
On Table~\ref{tab:res2} we show T(IoU), which measures the proportion of test images with IoU $\geq \kappa$, with $\kappa \in \{0.3,0.5,0.6\}$.
Figure~\ref{fig:samples} shows visual examples of classification and detection results produced by our approach.

\begin{table}[]
\centering
\caption{
Comparison on classification results of state-of-the-art methods on ChestX-ray14.
}
\scalebox{0.9}{
\begin{tabular}{l|c|c|c|c|c}
\hline
Label & Wang \textit{et al.}  & Li \textit{et al.}  & CRAL  & Ma \textit{et al.}  & Ours \\
& \cite{wang_chestx-ray8_2017} & \cite{li_thoracic_2018} & \cite{guan_multi-label_2018} & \cite{ma_multi-label_2019} & \\ \hline
Atelectasis  & ~0.700~ & ~0.729~ & \textbf{~0.781~} & ~0.777~ & ~0.775~ \\
Cardiomegaly  & ~0.810~ & ~0.846~ & ~0.883~ & \textbf{~0.894~} & ~0.881~ \\
Effusion  & ~0.759~ & ~0.781~ & \textbf{~0.831~} & ~0.829~ & ~\textbf{0.831}~ \\
Infiltration  & ~0.661~ & ~0.673~ & ~\textbf{0.697}~ & ~0.696~ & ~0.695~ \\
Mass  & ~0.693~ & ~0.743~ & ~0.830~ & \textbf{~0.838~} & ~0.826~ \\
Nodule  & ~0.669~ & ~0.758~ & ~0.764~ & ~0.771~ & \textbf{~0.789~} \\
Pneumonia  & ~0.658~ & ~0.633~ & ~0.725~ & ~0.722~ & \textbf{~0.741~} \\
Pneumothorax  & ~0.799~ & ~0.793~ & ~0.866~ & ~0.862~ & \textbf{~0.879~} \\
Consolidation  & ~0.703~ & ~0.720~ & \textbf{~0.758~} & ~0.750~ & ~0.747~ \\
Edema  & ~0.805~ & ~0.710~ & \textbf{~0.853~} & ~0.846~ & ~0.846~ \\
Emphysema  & ~0.833~ & ~0.751~ & ~0.911~ & ~0.908~ & \textbf{~0.936~} \\
Fibrosis  & ~0.786~ & ~0.761~ & ~0.826~ & ~0.827~ & \textbf{~0.833~} \\
Pleural thickening  & ~0.684~ & ~0.730~ & ~0.780~ & ~0.779~ & \textbf{~0.793~} \\
Hernia  & ~0.872~ & ~0.668~ & ~0.918~ & \textbf{~0.934~} & ~0.917~ \\ \hline
Mean  & ~0.745~ & ~0.739~ & ~0.816~ & ~0.817~ & \textbf{~0.821~} \\ \hline
\end{tabular}
}
\label{tab:res1}
\end{table}

\begin{table}[]
\centering
\caption{
Comparison of localisation measures: average IoU and cDice~\cite{yao_weakly_2018} (Pne1 represents Pneumonia and Pne2 Pneumothorax).
Our methods Sal, Det and Mix are describe as follows:
1) Sal, uses saliency map $\mathbf{x}_S$;
2) Det, uses saliency map $\mathbf{x}_{RPN}$;
and 3) Mix, uses saliency map $\tilde{\mathbf{x}}_{FS}$.
We show in bold the best results within a 0.005 confidence.
}
\scalebox{0.9}{
\begin{tabular}{l|l|c|c|c|c|c|c|c|c|c}
\hline
Metric~ & Method & Atel & Card & Effu & Infi & Mass & Nodu & Pne1 & Pne2 & Mean \\ \hline
IoU & Sal & 0.201 & 0.558 & 0.208 & 0.300 & 0.230 & 0.080 & 0.335 & \textbf{0.144} & 0.257 \\
 & Det & ~0.216~ & ~\textbf{0.663}~ & ~\textbf{0.221}~ & ~0.322~ & ~0.216~ & ~0.081~ & ~0.327~ & ~0.128~ & ~0.272~ \\
 & Mix & \textbf{0.240} & ~\textbf{0.662} & \textbf{0.226} & \textbf{0.343} & \textbf{0.240} & \textbf{0.092} & \textbf{0.346} & 0.133 & \textbf{0.285} \\ \hline
cDice & \cite{yao_weakly_2018} & 0.204 & 0.180 & 0.293 & 0.325 & 0.202 & \textbf{0.295} & 0.112 & 0.039 & 0.206 \\
 & Sal & 0.296 & \textbf{0.737} & 0.333 & 0.374 & 0.294 & 0.059 & 0.424 & \textbf{0.222} & 0.342 \\
 & Det & 0.376 & 0.590 & \textbf{0.363} & \textbf{0.449} & 0.361 & 0.143 & \textbf{0.494} & 0.200 & \textbf{0.372} \\
 & Mix & \textbf{0.403} & 0.500 & 0.355 & 0.431 & \textbf{0.403} & 0.181 & \textbf{0.493} & 0.190 & \textbf{0.370} \\ \hline
\end{tabular}
}
\label{tab:comparison}
\end{table}

\begin{table}[]
\centering
\caption{
Comparison of localisation accuracy using IoU (Pne1 represents Pneumonia and Pne2 Pneumothorax).
}
\scalebox{0.9}{
\begin{tabular}{c|l|c|c|c|c|c|c|c|c|c}
 \hline
T(IoU) & Model~ & Atel & Card & Effu & Infi & Mass & Nodu & Pne1 & Pne2 & Mean \\ \hline
0.3 & \cite{wang_chestx-ray8_2017} & ~0.24~ & ~0.46~ & ~0.30~ & ~0.28~ & ~0.15~ & ~\textbf{0.04}~ & ~0.17~ & ~0.13~ & ~0.22~ \\
 & Ours & \textbf{0.37} & \textbf{0.99} & \textbf{0.37} & \textbf{0.54} & \textbf{0.35} & \textbf{0.04} & \textbf{0.60} & \textbf{0.21} & \textbf{0.43} \\ \hline
0.5 & \cite{wang_chestx-ray8_2017} & 0.05 & 0.18 & \textbf{0.11} & 0.06 & 0.01 & \textbf{0.01} & 0.03 & 0.03 & 0.06 \\
 & Ours & \textbf{0.11} & \textbf{0.92} & 0.05 & \textbf{0.30} & \textbf{0.13} & 0.00 & \textbf{0.27} & \textbf{0.06} & \textbf{0.23} \\ \hline
0.6 & \cite{wang_chestx-ray8_2017} & 0.02 & 0.08 & \textbf{0.05} & 0.02 & 0.00 & \textbf{0.01} & 0.02 & \textbf{0.03} & 0.03 \\
 & Ours & \textbf{0.04} & \textbf{0.73} & 0.01 & \textbf{0.20} & \textbf{0.05} & 0.00 & \textbf{0.18} & 0.01 & \textbf{0.15} \\ \hline
\end{tabular}
}
\label{tab:res2}
\end{table}

\section{Discussion and Conclusion}
\label{sec:disc}

Experimental results in Tables~\ref{tab:res1}~and~\ref{tab:res2} show that our proposed method establishes the new state-of-the-art performance results for the problem of disease classification and weakly supervised localisation in the ChestX-ray14 data set.
Compared to previous methods, our classification model benefits from using a feature pyramid network~\cite{lin_feature_2017} pipeline that considers features maps at several scales to account for the size variation of the pathologies.
In terms of localisation results, our method shows superior results compared with previous approaches -- we argue that this happens because methods based on saliency maps~\cite{wang_chestx-ray8_2017,yao_weakly_2018} suffer from the low resolution of intermediate feature maps. Similarly to classification, we alleviate this problem using FPN~\cite{lin_feature_2017} that produces the initial saliency map at higher resolution. 
We also observe that the saliency map tends to include areas bigger than the actual targeted area. Thus, we believe that saliency maps alone are not suited to obtain good localisation predictions. By including individual region proposals during training, our method is able to focus on different regions of the input separately and effectively refine localisation results.

In this paper, we proposed a new model for disease classification and weakly supervised localisation from chest X-ray images. Our model produces the disease classification using a saliency map that indicates the relevant regions for the classification. This localisation information is then refined using the  straight-through Gumbel-Softmax estimator to discretely sample region proposals, allowing the method to refine initially proposed regions in an end-to-end training set-up. 
Future work will focus on simultaneously improving the classification performance and refining detection results by modelling interactions between potential multiple regions of interest from the same image.

%
%
\bibliographystyle{splncs04}
\bibliography{zotero.bib}
%

\newpage
\beginsupplement
\section{Supplementary Material}

\begin{figure}[b!]
\centering
\includegraphics[width=0.99\linewidth]{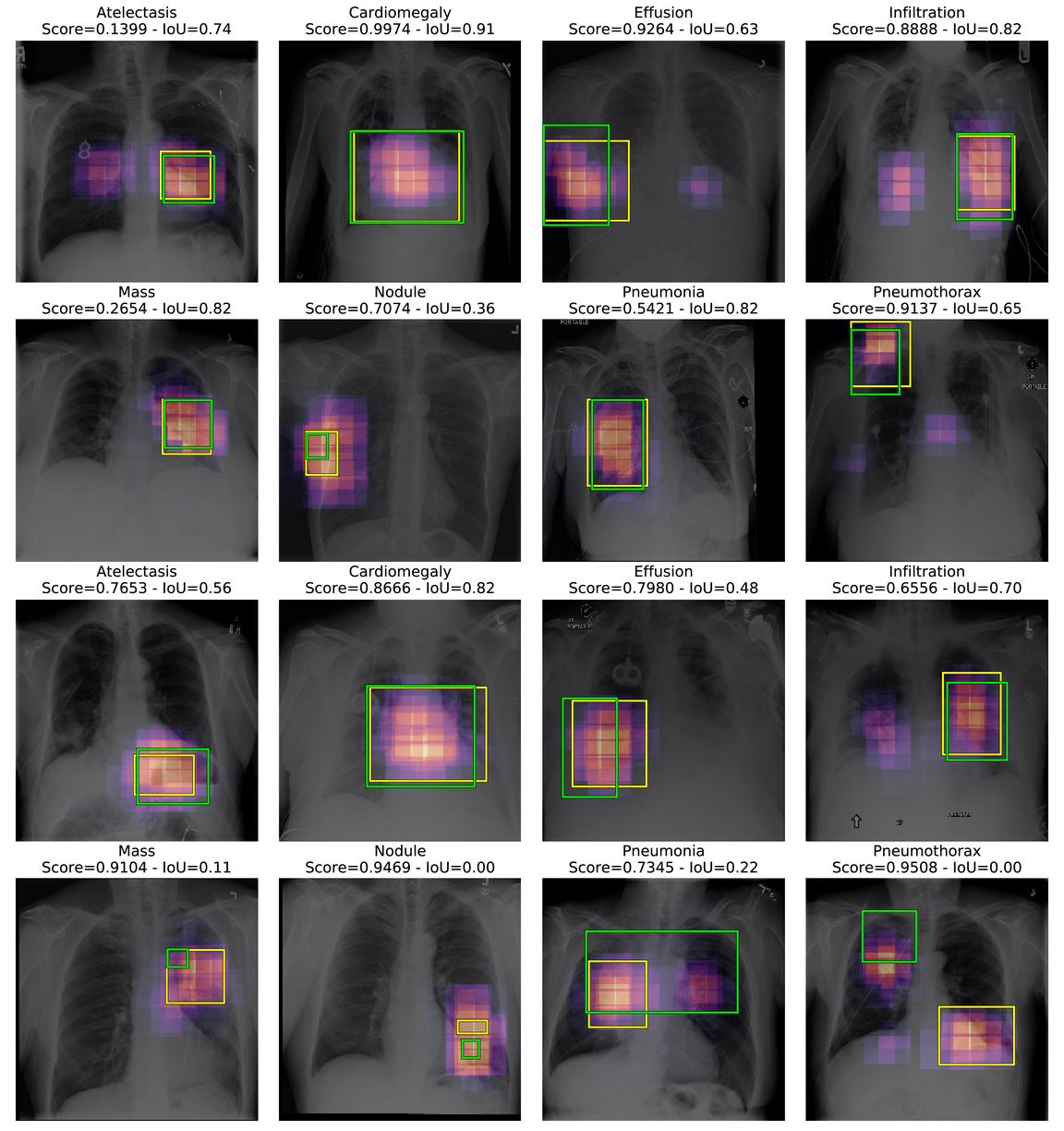}
\caption{Examples of classification and detection results for all classes.
We show some failure cases on the last row.
Predicted bounding boxes are shown in yellow and ground truth in green.}
\label{fig:extra1}
\end{figure}

\begin{figure}[t]
\centering
\includegraphics[width=1.00\linewidth]{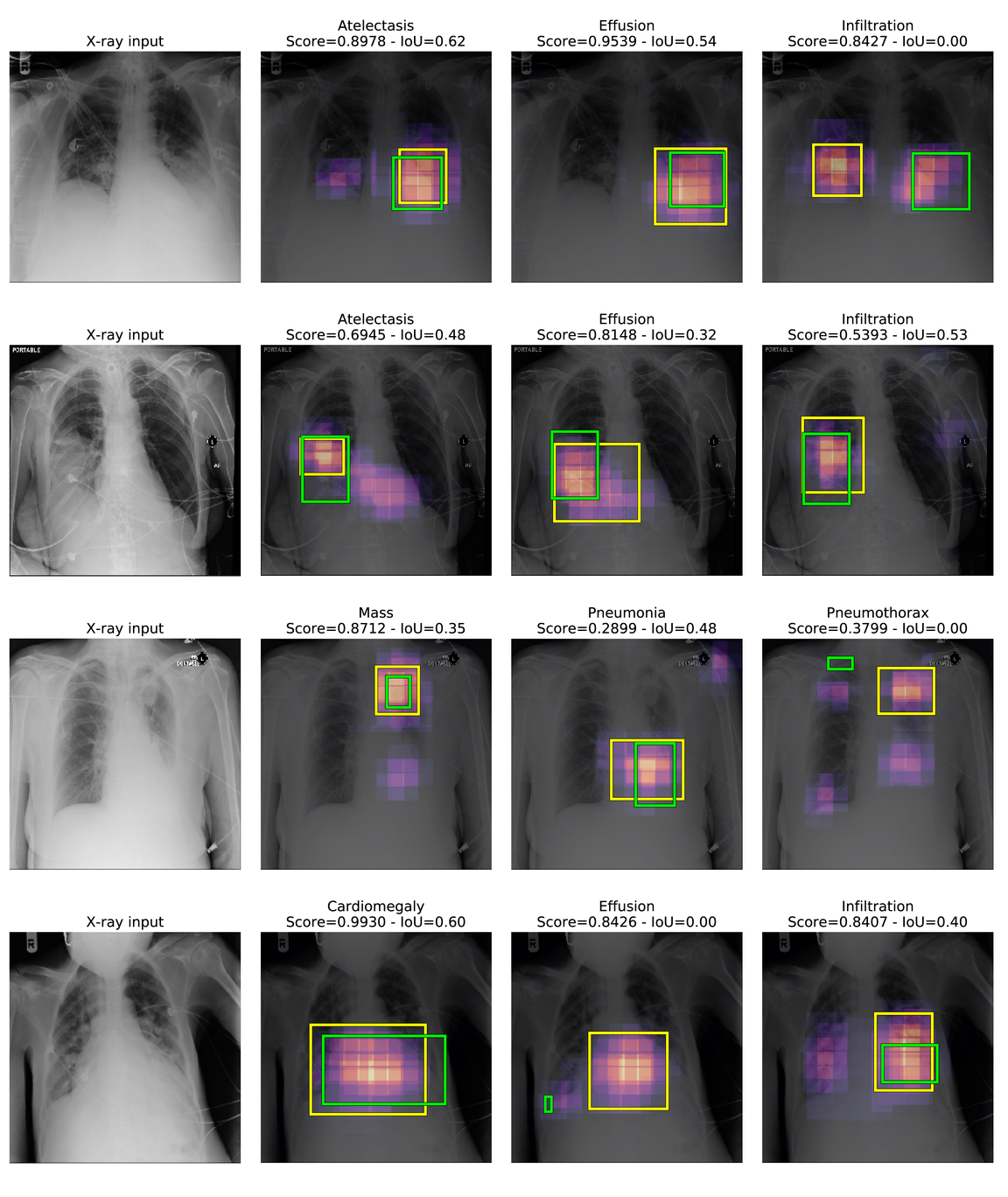}
\caption{
Examples of results on multi-class samples, each row shows a single case and the classes shown correspond to the ground truth.
Predicted bounding boxes are shown in yellow and ground truth in green.}
\label{fig:extra2}
\end{figure}

\end{document}